\documentclass[letterpaper,10pt, conference]{ieeeconf}
\IEEEoverridecommandlockouts      
\usepackage{graphics} 
\usepackage{graphicx} %

\usepackage{hyperref} 
\hypersetup{
    colorlinks=true,
    linkcolor=blue,
    filecolor=magenta,      
    urlcolor=blue,
    pdftitle={Overleaf Example},
    pdfpagemode=FullScreen,
}

\usepackage{amsmath} 
\usepackage{amssymb}  

\usepackage{gensymb}

\usepackage{array} 

\usepackage{dirtytalk} 

\usepackage{comment}

\usepackage{algorithm} 
\usepackage{algpseudocode} 

\usepackage{subcaption} 

\usepackage{titlesec}
\usepackage{tikz}
\usetikzlibrary{calc}

\usepackage{wrapfig}

\usepackage{mathtools}

\usepackage{cite}
\usepackage{multirow}

\let\labelindent\relax
\usepackage{enumitem}
\usepackage{soul}

\usepackage[font=footnotesize]{caption}

\graphicspath{{../figures/}}

\title{\LARGE \bf Dynamic Gap: Safe Gap-based Navigation in Dynamic Environments}

\author{{Max Asselmeier$^{1}$, Dhruv Ahuja$^{2}$, Abdel Zaro$^{3}$, Ahmad Abuaish$^{1}$, Ye Zhao$^{1}$, and Patricio A. Vela$^{1}$}
\thanks{$^{1}$M. Asselmeier, A. Abuaish, Y. Zhao, and P.A. Vela are with the Institute for Robotics and Intelligent Machines, Georgia Institute of Technology, Atlanta, GA 30308, USA.
		{\tt\small mass@gatech.edu}}%
\thanks{$^{2}$D. Ahuja is with the School of Electrical and Computer Engineering, Georgia Institute of Technology, Atlanta, GA 30308, USA.}
\thanks{$^{3}$A. Zaro is with the Department of Mechanical Engineering, University of California, Berkeley, Berkeley, CA, 94720, USA.}
\thanks{The work of Max Asselmeier is supported by the National Science Foundation Graduate Research Fellowship under Grant No. DGE-2039655. Any opinion, findings, and conclusions or recommendations expressed in this material are those of the authors(s) and do not necessarily reflect the views of the National Science Foundation.}}%

\begin{document}

\maketitle
\thispagestyle{empty}
\pagestyle{empty}


\begin{abstract}
This paper extends the family of gap-based local planners to unknown dynamic environments through generating provable collision-free properties for hierarchical navigation systems. Existing perception-informed local planners that operate in dynamic environments rely on emergent or empirical robustness for collision avoidance as opposed to performing formal analysis of dynamic obstacles. In addition to this, the obstacle tracking that is performed in these existent planners is often achieved with respect to a global inertial frame, subjecting such tracking estimates to transformation errors from odometry drift. The proposed local planner,  \textit{dynamic gap}, shifts the tracking paradigm to modeling how the free space, represented as gaps, evolves over time. Gap crossing and closing conditions are developed to aid in determining the feasibility of passage through gaps, and a breadth of simulation benchmarking is performed against other navigation planners in the literature where the proposed dynamic gap planner achieves the highest success rate out of all planners tested in all environments.
\end{abstract}



\section{Introduction} \label{sec:introduction}

Collision avoidance is of utmost importance for safe robot navigation. This task is typically handled by a local planner which utilizes sensory information to evade obstacles. One family of local planners is the gap-based planner \cite{mujahad_closest_2010}, which identifies passable regions, or “gaps”, and synthesizes motion commands through them. With this emphasis on free space, gap-based planners are an approach based on the \textit{affordances} of the environment \cite{gibson_ecological_2014}, and they have shown great promise with capabilities of respecting dynamic, visual, and geometric constraints \cite{sezer_novel_2012, mujahed_admissible_2018, mujahed_safe_2013, mujahed_new_2016, mujahed_tangential_2013} as well as generating provably collision-free trajectories \cite{xu_potential_2021}.

Despite this success, gap-based planners have only been extended to handling dynamic obstacle avoidance very recently \cite{contarli_predictive_2024}, a challenge that accurately reflects the unknown, changing environments of the real world. Reactive planners designed for static environments are often deployed in dynamic environments, relying on sufficient runtimes to adapt to the evolving environment. However, explicitly accounting for dynamic obstacles during planning allows for more assured collision avoidance behaviors.

\begin{figure}[t]
  \centering
  \includegraphics[width=0.48\textwidth]{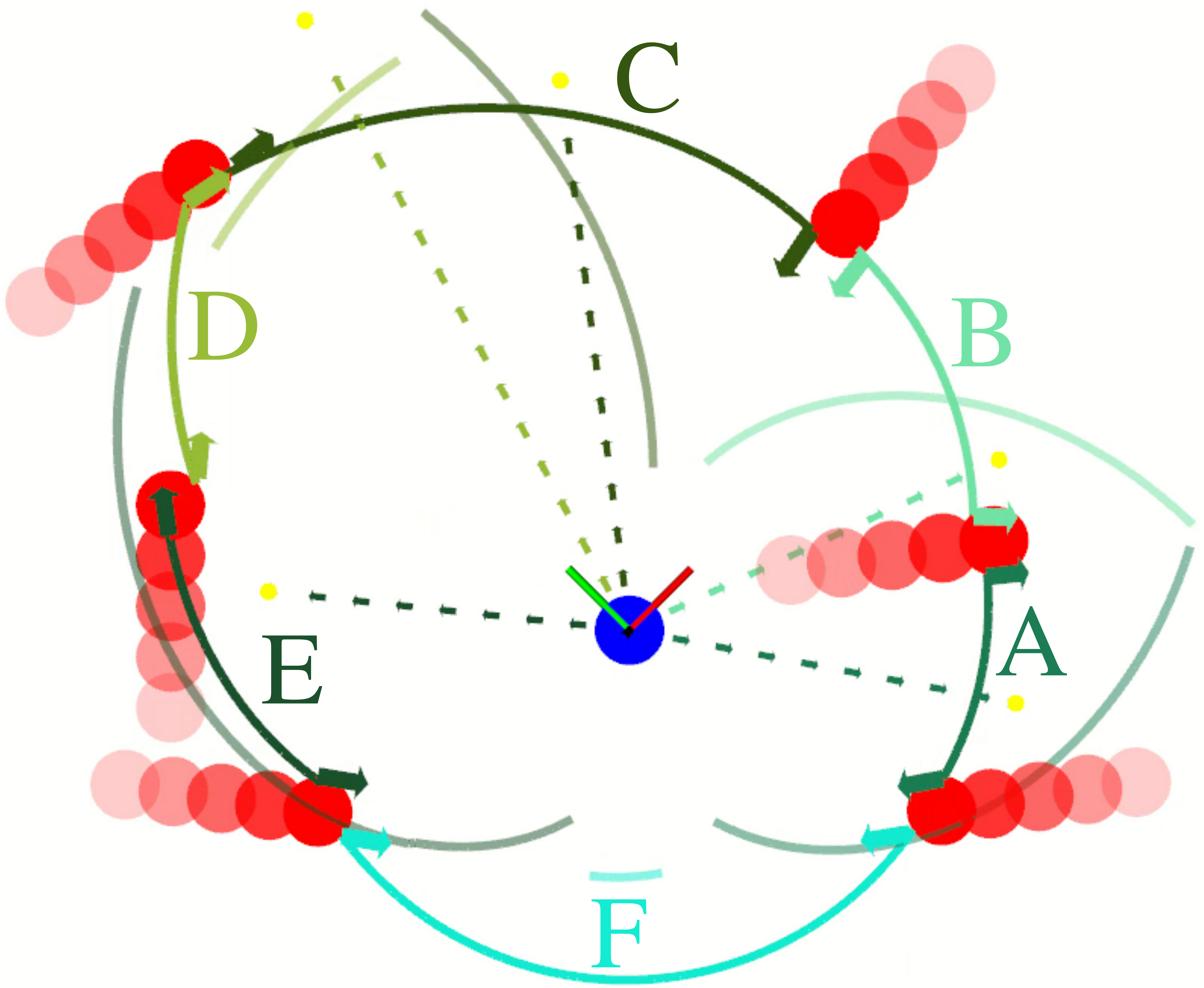}
  \caption{ Visualization of gaps and trajectories generated by dynamic gap. The central blue circle depicts the ego-robot while the red circle depict dynamic agents. The bold colored arcs labeled A-F are the instantaneous set of detected gaps and the transparent arcs show the predicted gaps obtained by propagating the gap dynamics models, shown as arrows, forward in time. Dashed lines are the candidate trajectories synthesized towards the gap goals shown in yellow. Gap F is predicted to close before the ego-robot can pass through, so it is deemed infeasible and not used during trajectory synthesis.}
  \label{fig:dgap_vis}
  \vspace{-0.3in}
\end{figure}

This paper details an extension to prior work on the \textit{potential gap} planner \cite{xu_potential_2021} involving augmentations to the planning framework to extend formal guarantees for safe navigation to dynamic environments under particular assumptions. Additional modules are also included to enable a more robust translation of planning success to non-ideal environments. This extension is referred to hereafter as \textit{dynamic gap} and can be visualized in Figure \ref{fig:dgap_vis}. Egocentric free space tracking is integrated to predict how gaps will evolve over time, and guidance law-based policies are employed to synthesize collision-free trajectories that pass through moving gaps under ideal conditions. Additional modules are then applied to robustify planning performance in non-ideal conditions.


Our main contributions are summarized as below:
\begin{enumerate}   
    \item Proposing an alternative tracking paradigm which revolves around the tracking and prediction of free space
    \item Adapting geometric and kinematic rules from guidance laws to the realm of gap-based planning to aid in dynamic gap propagation and feasibility analysis
    \item Providing a proof of collision-free dynamic gap passage under ideal conditions along with supplementary safety modules for non-ideal conditions
    \item Benchmarking against state-of-the-art local planners in the Arena-Rosnav simulation environment
\end{enumerate}

This planner is open-sourced\footnote{\href{https://github.com/ivaROS/DynamicGap}{https://github.com/ivaROS/DynamicGap}} within the Arena-Rosnav benchmarking environment \cite{kastner_arena-rosnav_2021, kastner_arena-bench_2022, kastner_arena-rosnav_2023, kastner_arena_2024} and available to test. 

\section{Related Work} \label{sec:related_work}

\subsection{Perception Space and Gap-based Navigation} \label{sec:perception_space_and_gap_based_navigation}
Most motion planners \cite{lavalle_rapidly-exploring_1998, rosmann_timed-elastic-bands_2015, connell_dynamic_2017} opt to plan using Cartesian world frame environment representations such as cost maps or voxel grids. These approaches contrast with the perception-space approach to planning which involves keeping sensory input in its raw egocentric form to take advantage of the computational benefits that come with foregoing intensive data processing. All local planning steps downstream are then cast as ego-centric decision making processes. 

Gap-based planners \cite{sezer_novel_2012, mujahed_admissible_2018, mujahed_safe_2013, mujahed_new_2016, feng_gpf-bg_2023, mujahed_tangential_2013, mujahed_admissible_2017, mujahad_closest_2010, ullah_fnug_2022, demir_improved_2017, contarli_predictive_2024, feng_safer_2023, chen_safe_2022} are an example of perception space-based planners. They detect regions of collision-free space defined by either leading or trailing edges of obstacles which can be viewed as an alternative way of discretizing the environment, here in the egocentric polar space as opposed to common methods such as occupancy maps. These planners then synthesize local trajectories or reactive control inputs  through these gaps. Some attention has been given to gaps in dynamic environments \cite{chen_safe_2022, contarli_predictive_2024}, but these methods do not develop their theory through a perception-informed approach, instead opting to use ground truth agent pose information. In the proposed work, explicit attention is paid towards how the dynamics of gaps must be ascertained from scan data in order to understand how the local gaps evolve over time.

\subsection{Guidance Laws} \label{sec:pursuit_guidance}

Guidance laws comprise a set of kinematic equations and feedback control laws that define collision course behavior between a pursuer and a target. While commonly affiliated with older forms of missile guidance, these laws have also seen use in many robotics applications \cite{noauthor_implementation_nodate, wellhausen_artplanner_2023, fiorini_motion_1998}.

Among the more established guidance laws, the two geometrical rules of pure pursuit and parallel navigation are the most popular. The pure pursuit rule, sometimes referred to as pursuit guidance, has the pursuer direct their velocity vector towards the target at all times, always keeping the target within the pursuer’s line of sight. Pure pursuit has seen a great deal of attention due to its simplicity \cite{bernhart_polygons_1959, bernhart_curves_1959, bruckstein_why_1993}, but this guidance law only leads to a collision if the pursuer is capable of traveling at a speed faster than that of the target. 

The parallel navigation, or constant bearing, rule \cite{rajasekhar_fuzzy_2000, ulybyshev_terminal_2005} has the pursuer direct their velocity vector such that the direction of the line of sight between the pursuer and target remains constant while the distance between them decreases. This geometrical rule is capable of yielding collision course conditions even if the pursuer is traveling slower than the target. Furthermore, for a non-maneuvering target, meaning a target that is not changing its speed nor its heading direction, parallel navigation is the optimal guidance law which yields a minimum intercept time.

\begin{figure*}[t!] \label{fig:info_flow}
    \centering
    \begin{tikzpicture}
          \hspace*{-0.7in}
        \node[] at ($(0, 0)$){{\scalebox{0.79} {\tikzstyle{block} = [draw, rectangle, text centered, thick,rounded corners=2pt,
                     minimum height=1.5em, minimum width=5em, inner sep=4pt]
\tikzstyle{contribBlock} = [draw, rectangle, text centered, ultra thick,
					 minimum height=1.5em, 
					 minimum width=5em, inner sep=4pt,
					 dash pattern=on 1pt off 2pt on 4pt off 2pt]
\tikzstyle{typical} = [fill=white!95!black]
\tikzstyle{reddish} = [draw=red,fill=white!95!red]
\tikzstyle{blueish} = [draw=blue,fill=white!95!blue]
\tikzstyle{greenish} = [draw=green!40!gray,fill=white!95!green]
\tikzstyle{longblock} = [draw,rectangle,text centered,thick,rounded corners=2pt,
                     minimum height=2em, minimum width=2em, inner sep=4pt]
\tikzstyle{contribLongblock} = [draw, rectangle, text centered, ultra thick,
					 minimum height=1.5em, 
					 minimum width=7.5em, inner sep=4pt,
					 dash pattern=on 1pt off 2pt on 4pt off 2pt]
\tikzstyle{largeBlock} = [draw, rectangle, thick,
                     minimum height=45em, minimum width=30em, inner sep=4pt]
\tikzstyle{smallBlock} = [draw, rectangle, text centered,
                     minimum height=2em, minimum width=4em]
\tikzstyle{dashedBlock} = [draw, dashed, rectangle, ultra thick, minimum height=7em, minimum width=55em, inner sep=4pt]
\tikzstyle{dashedWideBlock} = [draw, dashed, rectangle, ultra thick, minimum height=5em, minimum width=32.0em, inner sep=4pt]
\tikzstyle{dottedBlock} = [draw, rectangle, text centered, ultra thick,
					 minimum height=1.5em, 
					 minimum width=8em, inner sep=4pt,
					 dash pattern=on 1pt off 2pt on 4pt off 2pt] 
\tikzstyle{newtip} = [->, very thick]
\tikzstyle{notip} = [-, very thick]
\tikzstyle{bidir} = [<->, very thick]
\tikzstyle{newtip_dashed} = [->, very thick, dashed]

\begin{tikzpicture}[auto, inner sep=0pt, outer sep=0pt, >=latex]

\node[anchor=center](goal) at (0.0, 0.0) {Goal};

\node[block, typical, anchor=center] (worldmap)  at ($(goal.east) + (1.5, 0.0)$) {\centering World Map};

\node[longblock, typical, anchor=center, text width=8em] (global_planning) at ($(worldmap.east) + (4.0, 0.0)$) {\centering Global Planner};

\node[anchor=center](levels) at ($(global_planning.east) + (12.75, 1.0)$) {\sc Level};
\node[anchor=center, text width=4em](high) at ($(levels.south) + (0.25, -1.0)$) {\centering High};

\draw[newtip] ($(goal.east) + (0.1, 0.0)$) -- (worldmap.west);

\draw[newtip] (worldmap.east) -- (global_planning.west);


\node[block,reddish,anchor=center, text width=3em] (scan) at ($(goal.center) + (0.0, -2.25)$) {\centering Scan};

\node[longblock,reddish,anchor=center, text width=6em] (gap_det) at ($(scan.east) + (2.625, 0.0)$) {\centering Gap \\ Detection \\ (Sec. \ref{sec:gap_det_simp})};

\node[contribBlock,reddish,anchor=center, text width=6em] (gap_est) at ($(gap_det.east) + (2.0, 0.0)$) {\centering Gap \\ Estimation \\ (Sec. \ref{sec:gap_est})};

\node[contribBlock,blueish,anchor=center, text width=8em] (pursuit_guidance) at ($(gap_est.east) + (2.25, 0.0)$) {\centering Gap \\ Propagation (Sec. \ref{sec:gap_propagation})};

\node[contribBlock,blueish,anchor=center, text width=8em] (traj_gen) at ($(pursuit_guidance.east) + (2.375, 0.0)$) {\centering Gap Trajectory \\ Generation \\ (Sec. \ref{sec:traj_gen})};

\node[contribBlock,blueish,anchor=center, text width=10em] (traj_comp) at ($(traj_gen.east) + (2.625, 0.0)$) {\centering Trajectory Scoring \\ and Selection \\ (Sec \ref{sec:traj_score})};

\node[anchor=center, text width=4em](mid) at ($(high.south) + (0.0, -1.75)$) {\centering Mid};

\node[dashedBlock, draw=blue!70!black,anchor=center] (dynamic_gap) at (10.75, -2.125) {};

\node[anchor=center] (dynamic_gap_text) at ($(dynamic_gap.north) - (0, 0.25)$ ) {\centering \textbf{Dynamic Gap}};


\draw[newtip] (scan.east) -- (gap_det.west);

\draw[newtip] ($(scan.east) + (0.95, 0.0)$) -- (worldmap.south);

\draw[newtip] (gap_det.east) -- (gap_est.west);

\node[anchor=center] (gaps_t_first) at ($(gap_det.east) + (0.35, 0.35)$ ) {\centering $\mathcal{G}_t$};

\draw[newtip] (gap_est.east) -- (pursuit_guidance.west);

\node[anchor=center] (gaps_t_second) at ($(gap_est.east) + (0.35, 0.35)$ ) {\centering $\mathcal{G}_t$};

\draw[newtip] ($(gap_est.north) + (0.0, 0.5)$) -- (gap_est.north);
\node[anchor=center] (gaps_tmin1) at 
($(gap_est.north) + (0.5, 0.25)$ ) {\centering $\mathcal{G}_{t-1}$};

\draw[newtip] (gap_est.east) -- (pursuit_guidance.west);

\node[anchor=center] (gaps_t_feas) at ($(pursuit_guidance.east) + (0.40, 0.35)$ ) {\centering $\mathcal{G}_t^{\rm feas}$};

\draw[newtip] (pursuit_guidance.east) -- (traj_gen.west);

\draw[newtip] (traj_gen.east) -- (traj_comp.west);

\node[anchor=center] (trajs_t) at ($(traj_gen.east) + (0.35, 0.35)$ ) {\centering $\mathcal{T}_t$};

\draw[newtip] ($(traj_comp.north) + (0.0, 0.5)$) -- (traj_comp.north);

\node[anchor=center] (trajs_tmin1) at 
($(traj_comp.north) + (0.5, 0.375)$ ) {\centering $\tau_{t-1}^{\rm local}$};

\draw[newtip] (traj_gen.east) -- (traj_comp.west);


\node[anchor=center, text width=4em](robot_state) at ($(scan.center) + (0.0, -3.5)$) {\centering Robot State};

\node[longblock,greenish,anchor=center, text width=12em] (traj_contr) at ($(robot_state.east) + (6.0, 0.0)$) {\centering MPC Trajectory Tracking \\  (Sec. \ref{sec::mpc})};

\node[longblock,greenish,anchor=center, text width=10em] (proj_op) at ($(traj_contr.east) + (4.0, 0.0)$) {\centering Projection Operator \\ (Sec. \ref{sec:po})};

\node[longblock, typical, anchor=center, text width=5em] (robot) at ($(proj_op.east) + (2.5, 0.0)$) {Robot};

\node[anchor=center, text width=4em](low) at ($(mid.south) + (0.0, -3.25)$) {\centering Low};;

\node[dashedWideBlock, draw=green!70!black,anchor=center] (traj_tracker) at (9.625, -5.5) {};

\node[anchor=center] (traj_tracker_text) at ($(traj_tracker.north) - (0, 0.25)$ ) {\centering \textbf{Trajectory Tracker} };

\draw[newtip] (robot_state.east) -- (traj_contr.west);

\node[anchor=center] (v_cmd) at ($(traj_contr.east) + (1.25, 0.3)$ ) {\centering $\mathbf{v}_t^{\rm cmd}$};

\draw[newtip] (traj_contr.east) -- (proj_op.west);

\node[anchor=center] (v_cmd) at ($(proj_op.east) + (1.0, 0.3)$ ) {\centering $\mathbf{v}_t^{\rm safe}$};

\draw[newtip] (proj_op.east) -- ($(proj_op.east) + (1.5, 0.0)$);

\draw[notip] (traj_comp.south) -- ($(traj_comp.south) + (0.0, -1.25)$);

\draw[notip] ($(traj_comp.south) + (0.0, -1.25)$) -- ($(traj_comp.south) + (-11.525, -1.25)$);

\node[anchor=center] (traj_t_best) at 
($(traj_comp.south) + (-6.0, -1.0)$) {\centering $\tau_{t}^{\rm local}$};

\draw[newtip] ($(traj_comp.south) + (-11.525, -1.25)$) -- (traj_contr.north);

\draw[notip] (global_planning.east) -- ($(global_planning.east) + (5.7, 0.0)$);

\draw[newtip] ($(global_planning.east) + (5.7, 0.0)$) -- (traj_gen.north);

\node[anchor=center] (traj_t_best) at 
($(traj_gen.north) + (0.6, 1.0)$) {\centering $\tau_{t}^{\rm global}$};

\end{tikzpicture}}}};
    \end{tikzpicture}
    \caption{Overall workflow for the proposed navigation framework. Red blocks correspond to perceptual modules run at the rate of the laser scanner. Blue blocks are the core planning loop, and green blocks are the lower-level trajectory tracking routine. Dashed outlines represent core contributions.}
    \vspace{-0.25in}
\end{figure*}
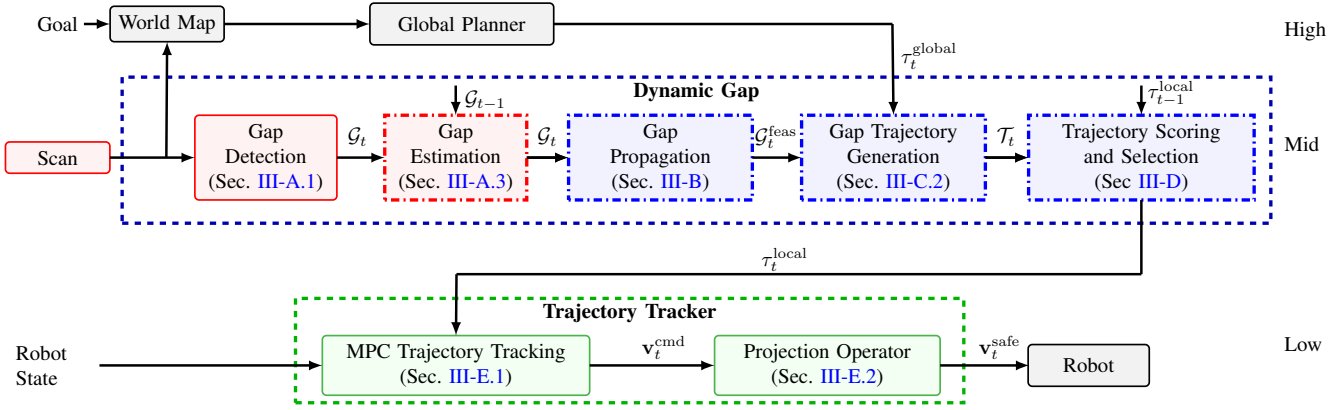

\section{Dynamic Gap Local Planning Module} \label{dynamic_gap_module}
\subsection{Gap Detection, Association, and Estimation} \label{sec:gap_det_assoc_est}
\subsubsection{Gap Detection and Simplification} \label{sec:gap_det_simp}

As input to the planner, we assume access to a 360$\degree$ laser scan $\mathcal{L}$. 
Gap detection involves the parsing of this scan $\mathcal{L}$ to obtain a set of detected gaps $\mathcal{G}_{\rm det}$ that describe the instantaneous free space of the local environment, further details regarding how this gap detection policy is defined can be found at \cite{xu_potential_2021}. Once this set of gaps has been extracted from the laser scan, an additional pass through $\mathcal{G}_{\rm det}$ is performed to remove any redundant gaps and potentially merge adjacent gaps together, yielding a set of simplified gaps $\mathcal{G}_{\rm simp}$.   


\subsubsection{Gap Association} \label{sec:gap_assoc}
The set of simplified gaps $\mathcal{G}_{\rm simp}$ captures the immediate free space, but we seek to understand how this free space will evolve across our local planning horizon. Therefore, additional steps must be taken to track gap points over time.

At this stage, each gap $g \in \mathcal{G}_{\rm simp}$ is characterized by a left gap point $\mathbf{p}_l = [x_l, y_l]^T$ and a right gap point $\mathbf{p}_r = [x_r, y_r]^T$. Association is performed on the set of points from the simplified gaps, 
\[
P_{\rm simp} = \{ \mathbf{p}_l^0, \mathbf{p}_r^0, ..., \mathbf{p}_l^N, \mathbf{p}_r^N \}.
\]
The association step is represented as a rectangular assignment problem, where the cost is equivalent to the distance between points across consecutive steps, $P_{\rm simp}^{t-1}$ and $P_{\rm simp}^{t}$. This assignment problem is then solved with the Hungarian Algorithm \cite{kuhn_hungarian_1955}, producing a minimum distance mapping between $P_{\rm simp}^{t-1}$ and $P_{\rm simp}^{ t}$. If the distance between two associated points exceeds a threshold $\tau_{assoc}$, then that point-to-point association is discarded.

\subsubsection{Gap Estimation} \label{sec:gap_est}
The point-to-point associations provide an insight into how the set of gap points are changing over time. This evolution is characterized with a second-order dynamics model with respect to the rotating ego-robot frame. 

The state vector is defined as:
\begin{equation}
    \textbf{X} = \begin{bmatrix}
                \mathbf{p}_{s/e} \\
                \mathbf{v}_{s/e} \\
                \end{bmatrix} = 
                \begin{bmatrix}
                \mathbf{p}_s - \mathbf{p}_e \\
                \mathbf{v}_s - \mathbf{v}_e
                \end{bmatrix},	
\end{equation}
where $\mathbf{p}_{s/e} \in \mathbb{R}^2$ and $\mathbf{v}_{s/e} \in \mathbf{R}^2$ represent the position and velocity of the gap side $s$ (left or right) relative to the ego-robot $e$, respectively. The system dynamics are then:
\begin{equation} \label{eq:rot_frame_dynamics}
    \dot{\textbf{X}} = \begin{bmatrix}
                \dot{\mathbf{p}}_{s/e} \\
                \dot{\mathbf{v}}_{s/e} \\
                \end{bmatrix} = 
                \begin{bmatrix}
                \mathbf{v}_{s/e} - \omega_e \times \mathbf{p}_{s/e}  \\
                \mathbf{a}_{s/e} - \omega_e \times \mathbf{v}_{s/e} \\
                \end{bmatrix},
\end{equation}
where $\omega_e$ represents the angular velocity of the ego-robot and $\mathbf{a}_{s/e}$ represents the linear acceleration of the gap side relative to the ego-robot. We make a constant velocity assumption on the gap points, which then simplifies Equation \ref{eq:rot_frame_dynamics} to
\begin{equation} \label{eq:rot_frame_dynamics_simp}
    \dot{\textbf{X}} = \begin{bmatrix}
                \dot{\mathbf{p}}_{s/e} \\
                \dot{\mathbf{v}}_{s/e} \\
                \end{bmatrix} = 
                \begin{bmatrix}
                \mathbf{v}_{s/e} - \omega_e \times \mathbf{p}_{s/e}  \\
                -\mathbf{a}_e - \omega_e \times \mathbf{v}_{s/e} \\
                \end{bmatrix}.
\end{equation}

This model is estimated with an extended Kalman filter given the nonlinear cross-product in Equation \ref{eq:rot_frame_dynamics_simp}. 

\subsection{Gap Propagation} \label{sec:gap_propagation}
Within static environments, gap feasibility solely depends on the geometric condition of whether or not the gap is wide enough to fit the robot. For dynamic environments, both geometric and kinematic considerations must be made to determine whether or not the robot can pass through the gap before the gap shuts. By pruning away infeasible gaps, this step not only conserves energy of the robot, but it also avoids potentially dangerous gaps through which it would be difficult, if not impossible, for the robot to pass. 


To understand the behavior of gaps over the time horizon for which planning will be performed, gap models (Eq. \ref{eq:rot_frame_dynamics_simp}) are integrated forward under the constant velocity assumption. In order to remove the ego-robot motion from the gap state and solely analyze the motion of the gap itself, the \textit{gap-only} dynamics are recovered from the prediction models by adding the ego-robot’s velocity $\mathbf{v}_e$ to the relative velocity estimate $\mathbf{v}_{s/e}$ to obtain the gap-only velocity $\mathbf{v}_s$. Gap propagation can terminate prematurely in two ways: 
\begin{enumerate}
    \item The gap closes, meaning if a gap that is shrinking over time reaches an angular span of $0$ rad, and
    \item The gap overlaps, meaning if a gap that is expanding over time passes beyond an angular span of $2\pi$ rad.
\end{enumerate} 

At each time step $t$ during integration, crossing and expanding conditions are checked for each gap. These conditions will now be further detailed.

\subsubsection{Crossing condition}
\begin{figure}[h!]
  \centering
  \includegraphics[width=0.99\linewidth]{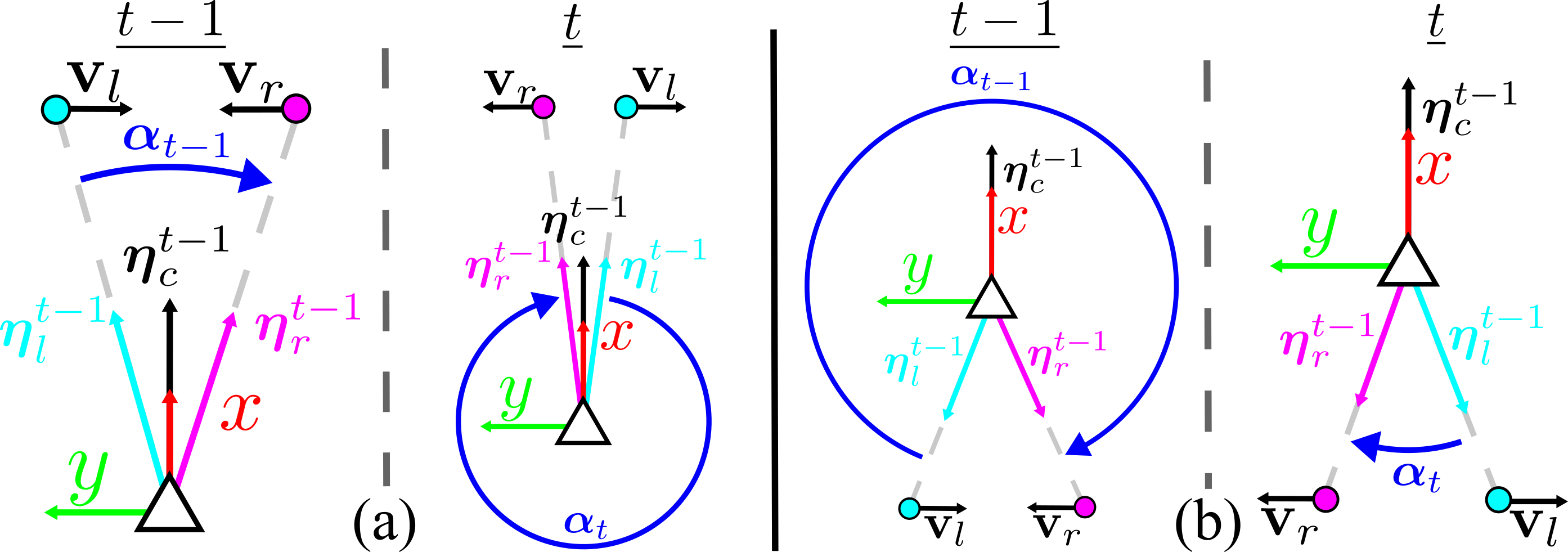}
  \caption{(a) Crossing condition for a gap, where between timesteps $t-1$ and $t$, the gap's angular span reaches $0$ rad. (b) Overlapping condition for a gap, where the gap's angular span reaches $2\pi$ rad. }
  \label{fig:crossing_and_overlapping_condition}
  \vspace{-0.125in}
\end{figure}

First, we define the unit norm bearing vector for the gap side $s$ at time $t$ as $\mathbf{\boldsymbol{\eta}}_s^t := \mathbf{p}_s^t / \| \mathbf{p}_s^t \| $. Then, we can calculate the clockwise angular gap span as 
\begin{equation}
    \alpha^t = \arctan(\frac{\boldsymbol{\eta}^t_l \cdot \boldsymbol{\eta}^t_r}{\boldsymbol{\eta}^t_l \times \boldsymbol{\eta}^t_r}),
\end{equation}
which we use to define the bearing of the gap center as $\beta_c^t := \beta_l^t - \alpha^t / 2$.
This in turn allows us to obtain the unit norm bearing vector for the gap center, $\boldsymbol{\eta}_c^t = [ \cos(\beta_c^t), \sin(\beta_c^t)]^T$. The left and right gap points have crossed each other if the two following conditions are true:
\begin{equation}
(\boldsymbol{\eta}^t_l \cdot \boldsymbol{\eta}^{t-1}_c > 0) \wedge (\boldsymbol{\eta}^t_l \cdot \boldsymbol{\eta}^{t-1}_c > 0),
\end{equation}
meaning that the gap points form a convex polar arc, and
\begin{equation}
    \alpha^t > \pi,
\end{equation}
meaning that the clockwise angular span from the left gap point to the right gap point exceeds $\pi$. This scenario is visualized in part (a) of Figure \ref{fig:crossing_and_overlapping_condition}.

\subsubsection{Overlapping condition}
A set of gap points have overlapped if the exact negation of the crossing conditions are met, meaning that
\begin{equation}
(\boldsymbol{\eta}^t_l \cdot \boldsymbol{\eta}^{t-1}_c < 0) \wedge (\boldsymbol{\eta}^t_l \cdot \boldsymbol{\eta}^{t-1}_c < 0),
\end{equation}
and
\begin{equation}
    \alpha^t < \pi.
\end{equation}
This scenario is visualized in part (b) of Figure \ref{fig:crossing_and_overlapping_condition}.
From this gap propagation step, we obtain a gap lifespan $t_f$. 

\subsection{Gap Feasibility Analysis} \label{sec:gap_feas_analysis}
Now that we have determined how long each gap will exist in the local environment for, we must determine if it is kinematically feasible for the ego-robot to pass through such gaps before they cease to exist.

\subsubsection{Guidance Law Analysis} \label{sec:pursuit_guidance_analysis}
In this section, we outline the guidance law-based trajectory generation scheme that dynamic gap employs in order to determine if a gap is kinematically feasible.

\begin{figure}[h!]
    \vspace{0.125in}
    \centering
    \includegraphics[width=0.80\linewidth]{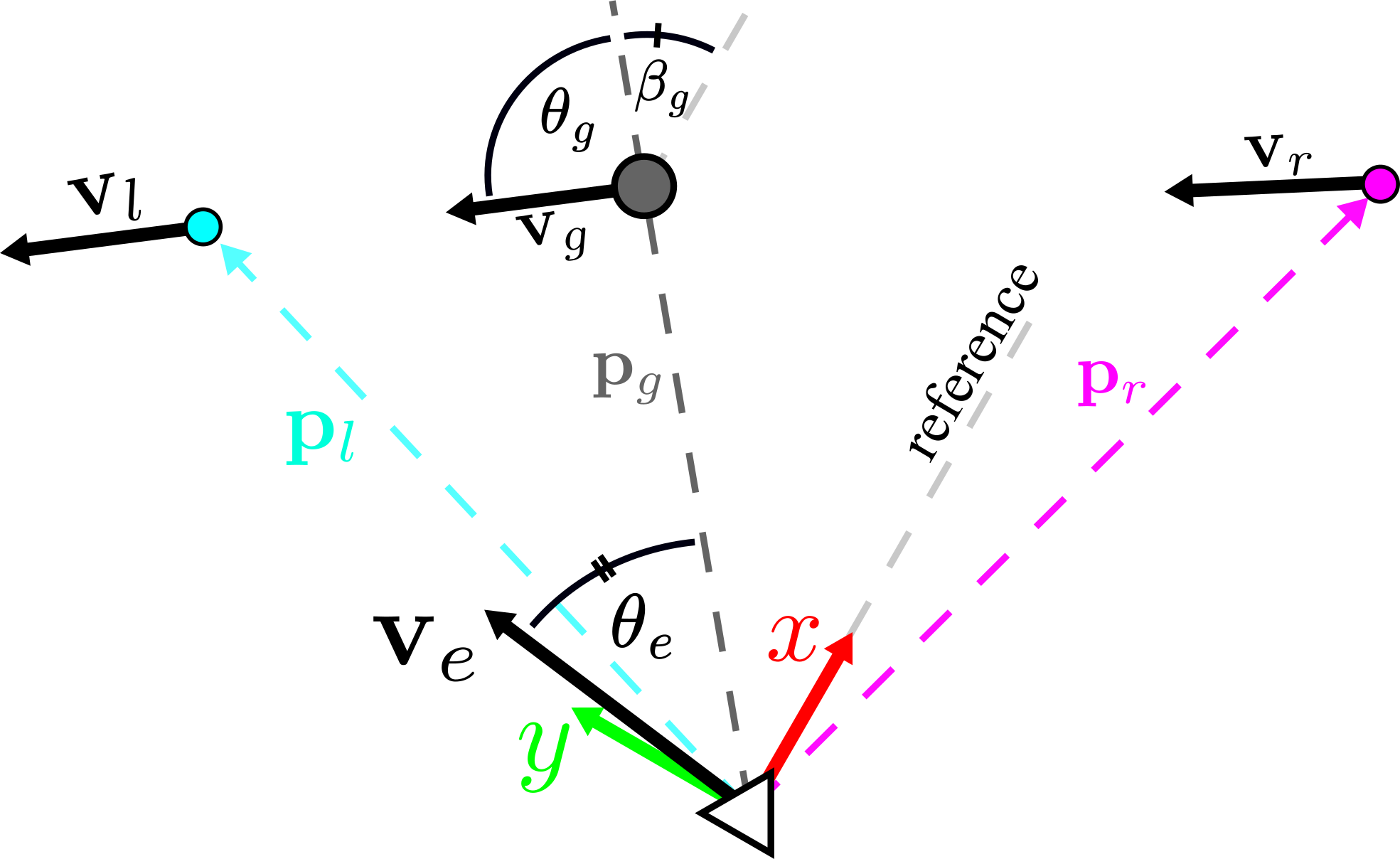}
    \caption{Diagram for guidance law notation.}
    \label{fig:pursuit_guidance_diagram}
    \vspace{-0.25in}
\end{figure}

We employ the parallel navigation policy \cite{shneydor_missile_1998}, for which we assume a constant gap goal point speed $v_g$ as well as a constant ego-robot speed $v_e$. This policy then dictates the bearing $\theta_e$ in which $v_e$ will be applied. With relation to Figure \ref{fig:pursuit_guidance_diagram}, the parallel navigation policy is defined as $\dot{\beta}_g = 0, \dot{r}_g < 0$, where
\begin{equation} \label{eq:rot_frame_dynamics}
    \begin{bmatrix}
                \dot{\beta}_g \\
                \dot{r}_g \\
                \end{bmatrix} = 
                \begin{bmatrix}
                \frac{v_g \sin(\theta_g) - v_e \sin(\theta_e)}{r_g} \\
                v_g \cos(\theta_G) - v_e \cos(\theta_e) \\
                \end{bmatrix},
\end{equation}
and $r_g := \| \mathbf{p}_g \|$. In order for $\dot{\beta}_g = 0$ to hold,
\begin{equation}
    v_g \sin(\theta_g) = v_e \sin(\theta_e).
\end{equation}
In order for $\dot{r}_g < 0$,
\begin{equation}
    v_e \cos(\theta_e) < v_g \cos(\theta_g).
\end{equation}
Therefore, for a constant speed ratio $K := v_e / v_g$, it follows that the gap goal position can be attained if
\begin{equation} \label{eq:theta_e}
    \sin(\theta_e) = \frac{\sin(\theta_g)}{K},
\end{equation}
and
\begin{equation}
    \cos(\theta_e) > \frac{\cos(\theta_g)}{K}.
\end{equation}
If these conditions can be met, then the ego-robot will intercept the goal position at the time
\begin{equation}
    t_{\rm intercept} = \frac{r_g^0}{v_g} \cdot \frac{1}{K \cdot \cos(\theta_e) - \cos(\theta_g)},
\end{equation}
where $r_g^0$ is equal to $r_g$ at $t=0$.
If these conditions can not be satisfied, then the given gap is deemed infeasible and discarded. If the conditions can be satisfied, but $t_f < t_{\rm intercept}$, meaning that the gap will cease to exist before the ego-robot can intercept the goal position, then the gap is also deemed infeasible and discarded.

\subsubsection{Collision-free Trajectory Generation} \label{sec:traj_gen}

If a gap is deemed feasible during the prior step, a collision-free trajectory can be synthesized by having the ego-robot direct its constant velocity along the bearing $\theta_e$ for at least $t_{\rm intercept}$ seconds. 



\subsubsection{Proof of Collision-Free Passage} \label{sec:proof} 
We aim to prove that performing the policy of parallel navigation towards the gap goal point $\mathbf{p}_g$ yields collision-free gap passage. Assumptions are as follows:
\begin{itemize}
    \item Ideal robot model: first-order, point-mass, constant velocity, holonomic system
    \item Constant velocity gap points
    \item Isolated gap: no other surrounding gaps will enter the gap in focus during the local time horizon
\end{itemize}
Let the gap goal point $\mathbf{p}_g$ and velocity $\mathbf{v}_g$ be defined as a convex combination of the left and right gap point states,
\begin{equation}
\begin{split}
    \mathbf{p}_g & = \kappa \mathbf{p}_l + (1 - \kappa) \mathbf{p}_r, \\
    \mathbf{v}_g & = \kappa \mathbf{v}_l + (1 - \kappa) \mathbf{v}_r, \hspace{0.5cm} \kappa \in [0, 1].
\end{split}
\end{equation}
It follows that
\begin{equation}
    \beta = \arctan(\mathbf{p}_g) = \arctan(\kappa \mathbf{p}_l + (1 - \kappa) \mathbf{p}_r).
\end{equation}
Without loss of generality (rotating the robot-centric frame to align with the center of the initial gap), $\beta_g \in [\beta_r, \beta_l]$ given that $\arctan$ is a monotonically increasing function. Following the same argument for
\begin{equation}
    \gamma_g = \arctan(\mathbf{v}_g) = \arctan(\kappa \mathbf{v}_l + (1 - \kappa) \mathbf{v}_r),
\end{equation}
it can be seen that $\gamma_g \in [\gamma_r, \gamma_l]$. Given that
\begin{equation}
    \gamma = \beta + \theta,
\end{equation}
it follows that $\theta_g \in [\theta_r, \theta_l]$.
From Equation \ref{eq:theta_e},
\begin{equation}
    \theta_e = \arcsin{( \frac{\sin{\theta_g}}{K})},
\end{equation}
and given that $\arcsin$ is also a monotonically increasing function, this means that $\theta_{e / g} \in [ \theta_{e / r}, \theta_{e / l}]$ where $\theta_{e / g}, \theta_{e / r}, \theta_{e / l}$ are the bearings at which the ego-robot must direct its velocity at to intercept the gap goal, left gap point, and right gap point, respectively. This indicates that under the parallel navigation policy, the ego-robot will intercept the gap goal point between the left and right gap points, therefore performing collision-free gap passage. 

\subsection{Trajectory Scoring} \label{sec:traj_score}
Each feasible gap $g_i$ produces a trajectory $\tau_i$. In order to determine which trajectory to track, each trajectory is evaluated by an egocentric pose-wise cost based on proximity to local obstacles and a terminal pose cost based on proximity to a local waypoint along the global plan in order to encourage progress toward the global goal. 

The trajectory cost formulation is adapted from \cite{xu_potential_2021}, with one key different: pose-wise scoring requires a laser scan for each timestep $t$ along the trajectory. We do not have direct access to this information, so we propagate gaps forward in time and back out propagated scans in practice.

The highest-scoring candidate trajectory is passed on to compared against the currently executing trajectory to determine if a trajectory change should occur. 

\subsection{Trajectory Tracking}

Trajectories are generated for the ideal unit single-integrator system. However, the simulation environment, Arena-Rosnav, adopts the non-ideal double-integrator system. The input to the system is the velocity command with a constraint on the command rate, i.e., acceleration constraints. Although reference acceleration commands can be extracted from the trajectory to facilitate trajectory tracking using feedforward control architecture, MPC is employed to track the reference trajectory.

\subsubsection{MPC Trajectory Tracking} \label{sec::mpc}
For the double integrator system $\mathcal{D}(\mathbf{x}, \mathbf{u})$, the state and control are $\mathbf{x}=[p_x,p_y,v_x,v_y]^T$ and $\mathbf{u}=[a_x, a_y]^T$ respectively. The state and control are uniformly discretized with a time step $dt=0.5$ over a maximum of $N=10$ timesteps.

The MPC optimization program is given in \ref{eq:whole_to}. 
\begin{subequations} \label{eq:whole_to}
\begin{align} 
& &&  \| \mathbf{x}[N]_{\{1:2\}]} - \mathbf{p}[N]^{\text{des}} \|^2 _{Q_f} + &&&  \\
& \min_{\mathbf{X}, \mathbf{U}} && \sum_{k=0}^{N - 1} \Biggl( \| \mathbf{x}[k]_{\{1:2\}} - \mathbf{p}[k]^{\text{des}}  \|^2 _Q + \| \mathbf{u}[k] \|^2 _{R} \Biggr) &&& \nonumber \\
& \text{s.t.}  && &&& \nonumber \\
& \text{(Dynamics)} && \mathbf{x}[k+1] = \mathcal{D}(\mathbf{x}[k], \mathbf{u}[k])  &&& \label{eq:dyn} \\
&&& \forall k \in \{0, \cdots, N-1\} \nonumber\\
& \text{(Initial state)} && x[0] = x_0 \label{eq:init}\\
& \text{(ZBF)} && A \mathbf{x}[j] \leq b &&& \label{eq:zbf} \\
& \text{(Velocity)} && |\mathbf{x}[j]_{\{3,4\}}| \leq v_{max} \label{eq:state_limits} \\
& \text{(Acceleration)} && \left| u[j]\right| \leq a_{max} &&& \label{eq:acc_limits}\\
&&&\forall j \in \{0, \cdots, N\} \nonumber
\end{align}
\end{subequations}
where the decision variables $\mathbf{X}=[\mathbf{x}[0], \cdots, \mathbf{x}[N]]^T$ and $\mathbf{U}=[\mathbf{u}[0], \cdots, \mathbf{u}[N-1]]^T$, the desired poses at discrete time $k$ from the best trajectory $\tau^t_{\rm best}$ are denoted by $\mathbf{p}[k]^{des}\in \mathbb{R}^2$, and the reduced projection of the first two elements of state is denoted by $x[k]_{\{1,2\}}$. The zeroing barrier function (ZBF) constraint ensures that control commands lie within the interval defined in Section \ref{sec:proof} for collision-free gap passage.
The optimization problem (\ref{eq:whole_to}) is solved using quadratic programming on MATLAB with an average compute time of 6.7 msec, and the MATLAB-ROS Toolbox was then used to communicate the MPC output to the system.
\subsubsection{Projection Operator} \label{sec:po}
As a last resort safety filter to handle non-ideal circumstances including discrete time implementation and second-order dynamics, the proposed work also adapts the projection operator module from \cite{xu_potential_2021}.

\section{Experimental Results}
This planner is implemented as a C++ ROS node through the \texttt{move\_base} package. All benchmarks are run on a Dell Precision 3660 Tower with an Intel i9-12900K CPU with 24 cores. The planning loop can run at $\approx50$ Hz. 




\begin{figure}[h!]
    \centering
    \includegraphics[width=0.99\linewidth]{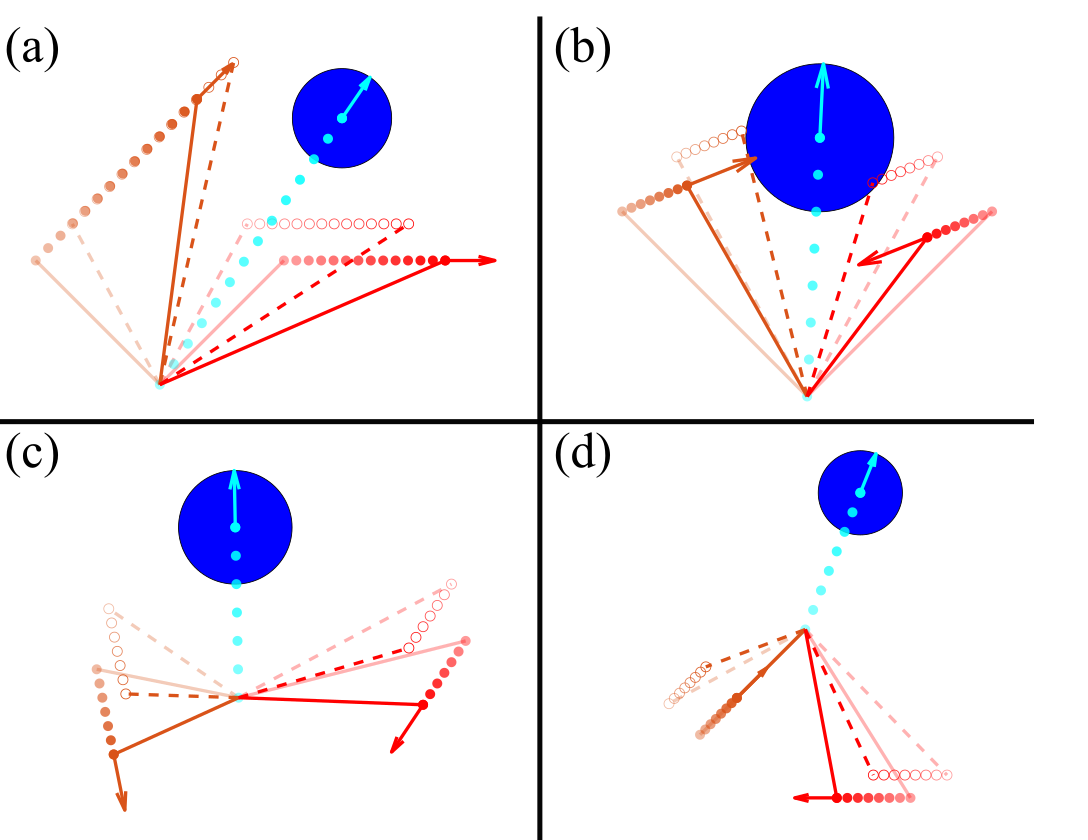}
    \caption{Visualization of four Monte Carlo variations of gaps from Experiment 1. Orange points and lines represent the left side of the gap while red points and lines represent the right side. Solid points and lines represent the original gap geometry whereas hollow points and dashed lines correspond to the inflated version of the gap. Transparent points represent positions at prior timesteps. The blue circle represents the robot along with its finite radius.}
    \label{fig:matlab_benchmark}
    \vspace{-0.25in}
\end{figure}

\begin{figure*}[t!]
    \vspace{0.1in}
    \centering
    \includegraphics[width=0.99\linewidth]{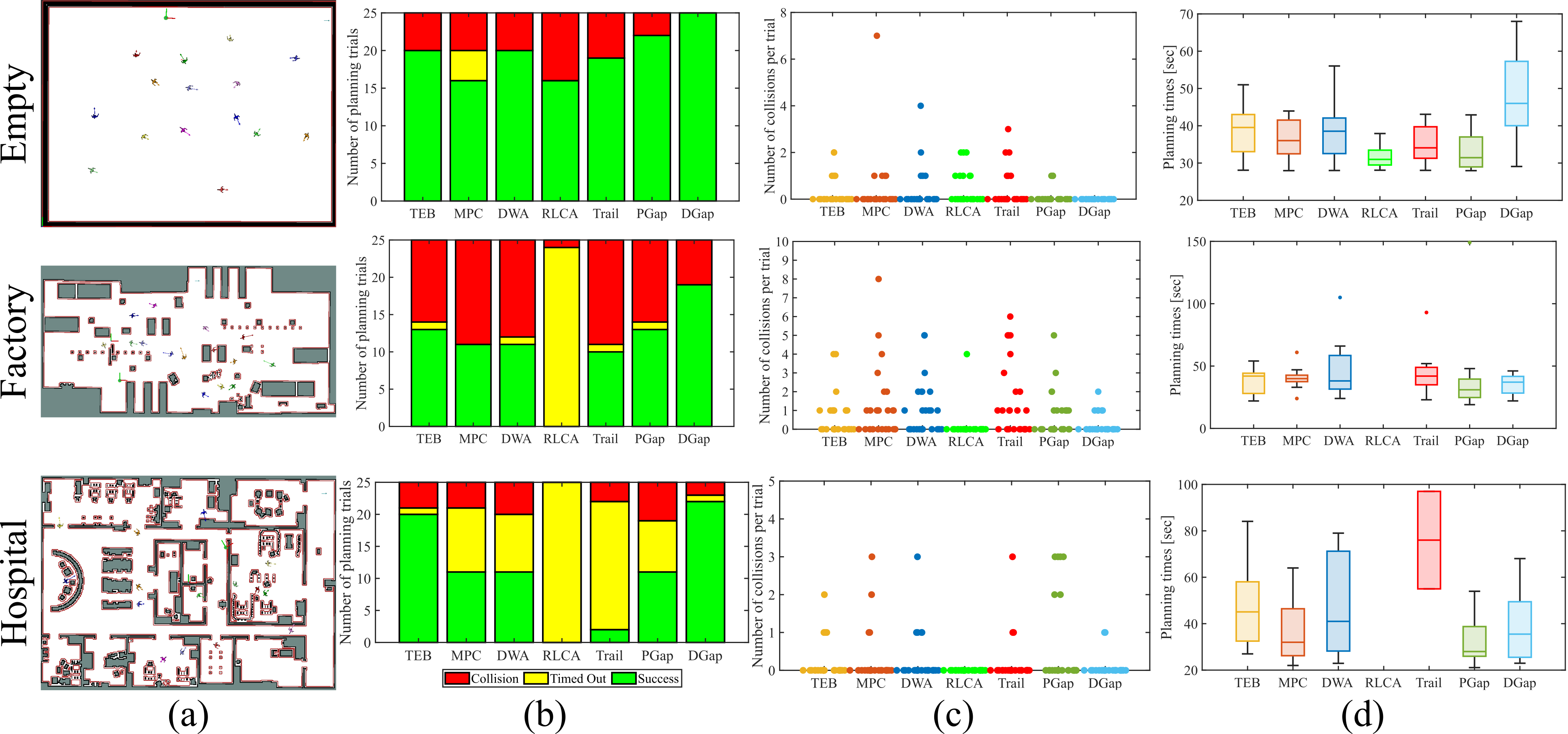}
    \caption{. Benchmarking results from Arena-Rosnav. (a) Visualization of map with ego-robot and agents, (b) planning trial outcomes, (c) distribution for number of collisions suffered per trial, and (d) average planning time per trial. }
    \label{fig:arena_benchmark}
    \vspace{-0.25in}
\end{figure*}

\subsection{Experiment 1: Assumption-adhering Environments}
First, we experimentally validate the proof of safe gap passage under the previously stated ideal conditions. To do so, a single gap is randomly generated and the parallel navigation policy is employed to generate a trajectory through the given gap. With the gap centered at the origin, left gap points are uniformly sampled from $\beta_l \in [\frac{\pi}{2}, \frac{3 \pi}{2}]$, $r_l \in [0.25, 1.0]$ m. Right gap points are uniformly sampled from $\beta_r \in [\frac{-\pi}{2}, \frac{\pi}{2}]$, $r_r \in [0.25, 1.0]$ m. Left and right gap point velocities are uniformly sampled from all directions with magnitudes within the range $[0.0, 1.0]$ m/s. A finite robot radius of $0.20$ m is also accounted for by artificially inflating gap points inwards during planning. A hand-selected set of such gaps are shown in Figure \ref{fig:matlab_benchmark}.

For this experiment, $10,000$ trials were run: $6,987$ trials end in the robot successfully passing through the gap, $2,668$ trials resulted in a kinematically infeasible gap due to the velocity limits of the robot, and for the remaining $345$ trials, the robot was unable to pass through the gap given its finite radius. No collisions occurred during any trials.

\subsection{Experiment 2: Assumption-violating Environments}
To contextualize dynamic gap's performance and understand the performance gap between ideal and non-ideal conditions, the planner is integrated into the Arena-Rosnav \cite{kastner_arena-rosnav_2021} benchmarking environment and compared against other state-of-the-art local planners. In this experiment, the classical motion planners DWA \cite{fox_dynamic_1997}, TEB \cite{rosmann_timed-elastic-bands_2015}, and MPC \cite{rosmann_online_2021} are tested along with the learned planners RLCA \cite{long_towards_2018} and Trail \cite{xie_drl-vo_2023}. Potential gap \cite{xu_potential_2021}, the predecessor to dynamic gap, is also tested. 

Three environments are used, shown in column (a) of Figure \ref{fig:arena_benchmark}, referred to from top to bottom as empty, factory, and hospital. With these environments, the authors aimed to build a gradient of increasing environment structure to evaluate each planner's ability to not only navigate dynamic obstacles, but also the static, non-trivial structure of these worlds such as corridors, rooms, and atria.

Within each environment, 15 dynamic agents are placed which maneuver between two manually defined waypoints within the environment. For each planner/environment combination, 25 trials are run, and results are reported by environment in Figure \ref{fig:arena_benchmark}. For all trials, planners are given five minutes to reach the goal.




\textit{Empty: } Given that this environment is solely comprised of a single room, these trials can be used to evaluate each planner's pure dynamic obstacle avoidance mechanisms. In this setting, the shortest path from start to goal is simply a straight line. Due to the lack of complexity in this setting, the variance in planning times is low, with most planners being able to reach the goal within $30-40$ seconds.

All other benchmarks perform fairly well in this setting, with success rates of $60-80\%$ across the board. Although, the set of classical planners, namely TEB and DWA, do perform slightly better than the learned planners. One interesting note from these trials was that for the RLCA benchmark, the planner tended to come to a stop when the robot encroached upon obstacles, relying on non-adversarial motions from the nearby agents in order to avoid collisions.


The ability of dynamic gap to predict the feasibility of local gaps before committing to them proved paramount in avoiding collisions in this environment. The proposed planner suffered no collisions over its trials, though this \say{patience} did result in longer planning times overall.



\textit{Factory: } The factory setting consists of one large room with many smaller isolated static obstacles positioned within it, resembling real-world artifacts such as tables or poles.

Here, the larger regions of free space allow for multiple agents to pass through the same part of the environment at one time. Because of this, more collisions are observed in this environment than the empty environment. General success rates for the benchmarks drops to roughly $50\%$, with the exception of RLCA failing every trial and dynamic gap registering an above average success rate of $72\%$. RLCA's performance can be explained in part by its neural network composition: inputs to this model are a sliding window of laser scan inputs, the relative goal position, and the robot's current state. While the scans provide a local environment representation, the model largely opts to drive in a straight line towards the goal. When structured obstacles such as walls are positioned between the robot and the goal, the planner struggles with maneuvering around such regions.




\textit{Hospital:} The hospital environment exhibits many smaller rooms connected through tighter passageways and corridors. While the prior environments allowed for more clearance en route to goal for planners, the primary failure mode observed in the hospital setting was the benchmarks exceeding the allotted time limit of five minutes rather than colliding with obstacles in the environment.

As can be seen in column (c), far fewer collisions are registered in this environment because the corridors and small rooms offered fewer opportunities for several agents to enter the same region and collide. Given that this environment is more sprawling, more variance is seen in planning times.

MPC and DWA stalled out on several trials when the robot was attempting to pass through thin entryways between rooms, eventually leading to a time out. Both TEB and dynamic gap performed well in this setting with success rates of $80\%$ and $88\%$, respectively. The Trail benchmark did exhibit smoother overall trajectories in this environment, though the planner preferred to take wider turns around corners, often times overshooting and running into walls. 

\section{Conclusion}
In this work, the perception-informed gap-based planning paradigm is extended to the case of the dynamic environment. In order to do so, gap dynamics models are estimated and propagated forward in time to evaluate the feasibility of gap passage, and guidance laws are employed to generate provably collision-free trajectories. Targeted modules including scan propagation, MPC trajectory tracking, and projection operator control modifications are employed to bridge the gap between single gap safety guarantees and multi-gap real world settings. In the future, the authors aim to extended these safety guarantees to the nonholonomic case and enable gap propagation to account for the potential creation of gaps over the local planning horizon.

\bibliographystyle{IEEEtran}
\bibliography{references}


\end{document}